\documentclass[sigconf]{acmart}

\usepackage{todonotes}
\usepackage{tabularx}
\usepackage{colortbl}
\usepackage{xcolor}
\usepackage{multirow}
\usepackage{enumitem}
\usepackage{caption}
\AtBeginDocument{%
  }

\setcopyright{acmcopyright}
\copyrightyear{2018}
\acmYear{2018}
\acmDOI{XXXXXXX.XXXXXXX}

\acmConference[Conference acronym 'XX]{Make sure to enter the correct
  conference title from your rights confirmation emai}{June 03--05,
  2018}{Woodstock, NY}
\acmPrice{15.00}
\acmISBN{978-1-4503-XXXX-X/18/06}

\begin{document}

\title{SciTweets - A Dataset and Annotation Framework for Detecting Scientific Online Discourse}


 \author{Salim Hafid}
\authornote{Both authors contributed equally to this research.}
 \email{salim.hafid@lirmm.fr}
\affiliation{%
  \institution{LIRMM, CNRS, University of Montpellier}
  \city{Montpellier}
  \country{France}
}

\author{Sebastian Schellhammer}
\authornotemark[1]
\email{sebastian.schellhammer@gesis.org}
\affiliation{%
  \institution{GESIS - Leibniz Institute for the Social Sciences}
 \city{Cologne}
  \country{Germany}}

\author{Sandra Bringay}
\affiliation{%
 \institution{LIRMM, CNRS, University of Montpellier}
  \city{Montpellier}
  \country{France}
  }
  
\author{Konstantin Todorov}
\affiliation{%
 \institution{LIRMM, CNRS, University of Montpellier}
  \city{Montpellier}
  \country{France}}

\author{Stefan Dietze}
\affiliation{%
  \institution{GESIS - Leibniz Institute for the Social Sciences}
 \city{Cologne}
   \institution{\& Heinrich-Heine-University}
  \city{Düsseldorf}
  \country{Germany}
  }
  
\renewcommand{\shortauthors}{Hafid, Schellhammer et al.}

\begin{abstract}
Scientific topics, claims and resources are increasingly debated as part of online discourse, where prominent examples include discourse related to COVID-19 or climate change. This has led to both significant societal impact and increased interest in scientific online discourse from various disciplines. For instance, communication studies aim at a deeper understanding of biases, quality or spreading pattern of scientific information whereas computational methods have been proposed to extract, classify or verify scientific claims using NLP and IR techniques. However, research across disciplines currently suffers from both a lack of robust definitions of the various forms of science-relatedness as well as appropriate ground truth data for distinguishing them. In this work, we contribute (a) an annotation framework and corresponding definitions for different forms of scientific relatedness of online discourse in Tweets, (b) an expert-annotated dataset of 1261 tweets obtained through our labeling framework reaching an average Fleiss Kappa \(\kappa\) of 0.63, (c) a multi-label classifier trained on our data able to detect science-relatedness with 89\% F1 and also able to detect distinct forms of scientific knowledge (claims, references). With this work we aim to lay the foundation for developing and evaluating robust methods for analysing science as part of large-scale online discourse. 
\end{abstract}

\vspace{-5em}



\maketitle

\section{Introduction}

Scientific topics, claims and resources are increasingly debated as part of societal discourse in online news and social media. Examples include the increased participation of journalists, policy makers, scientists, celebrities and the general public in scientific online discourse \citep{doi:10.1073/pnas.1805868115}, where Twitter in particular is used widely for discussing scientific insights (see examples in Table 1). Specifically for emerging topics such as COVID-19, an elevated role of preliminary scientific results beyond the traditional peer review system can be observed, for instance, as part of preprints, opinion pieces and informal utterances in scientific online debates \citep{doi:10.1177/09636625221077392}. 

\begin{table}[htpb]
    \caption{Examples (tweets 1 to 4) and Counterexamples (tweet 5) of scientific online discourse tweets}
    \label{tab:sod_examples}
    \begin{tabularx}{\columnwidth}{X}
        \toprule
        (1) Donating blood not only helps others, but reduces the rate of cancer and heart disease in the donor.\\
        \midrule
         (2) via @medical\_xpress A new in vitro (test tube) study, ""Dietary functional benefits of Bartlet http://t.co/Qv1C1GjQin \#UFO4UBlogHealth\\
        \midrule
        (3) How is @UChicagoIME shaping the future of science? Find out on April 6!\\ 
        \midrule
        (4) Study: Shifts in electricity generation spur net job growth, but coal jobs decline - via @DukeU http://t.co/AXGmKUPata\\
        \midrule
        (5) My father got COVID-19.\\
        \bottomrule
    \end{tabularx}
\end{table}
\vspace{-1em}
While it has been recognised that online discourse as observed in news and social Web platforms produces phenomena such as misinformation spread \citep{vosoughi} or reinforcement of biases \citep{quanticontroversy}  with widely assumed harmful effects for democratic societies \citep{AlcotttFakeNews}, misinformation on scientific topics such as COVID-19 or climate change has particularly detrimental effects on society and public health.

This has led to research into scientific online discourse across various disciplines. From a social sciences perspective, works measure the engagement with scientific publications on social media  \citep{carlson2020quantifying,haunschild2021investigating,diaz2019towards} or investigate the role of social media in facilitating the flow of scientific information \citep{banerjee2021democratization}. In science communication, research discusses implications of risk communication \citep{li2016tweeting} or the spreading pattern associated with preliminary scientific results and the diffusion of science through social networks \citep{moukarzel2020diffusing,walter2019scientific}, while research in cognitive and social psychology investigates the perceived trustworthiness of scientific online discourse \citep{doi:10.1126/sciadv.abd4563}. 

Methodological research at the intersection of NLP, information retrieval and machine learning is aimed at detecting, classifying or verifying (scientific) claims and discourse \citep{clefcheckthat2022,srba_monant_2022,smeros_sciclops_2021,hassan_claimbuster_2017,nakov_automated_2021}, and is a key facilitator for large-scale interdisciplinary analysis of science discourse. Prior works often focus on actual scholarly publications \citep{jansen2016extracting, Pinto2019WhatDR}, where the formality of language differs substantially from science claims in online news and social media, e.g., Twitter. 

Datasets are crucial to facilitate such research and were proposed with various definitions of science-relatedness that each are based on specific assumptions. Some works define scientific claims as claims expressing an aspect of one or more scientific entities \citep{wadden_fact_2020,smeros_sciclops_2021}, however with no robust definition of what a scientific entity is. Other works selected scientific claims by restricting the domain to one they deemed scientific (e.g COVID-19 \citep{saakyan_covid-fact_2021}, climate change \citep{diggelmann_climate-fever_2021}, or medicine \citep{srba_monant_2022}), where generalisability is limited.

Moreover, claims may be synthetically generated \citep{wright-etal-2022-generating,wadden_fact_2020} or ground truth data is constructed using simple heuristics exploiting keywords or referenced pay-level-domains (PLDs) based on narrow predefined dictionaries \citep{Scilens}, and predicates \citep{Pinto2019WhatDR}.
Generally, robust definitions of science-relatedness are lacking that distinguish between items that actually convey scientific knowledge, e.g., a science claim or a reference to a scientific resource, and other forms of science-relatedness, for instance, items stating a fact about a particular scientist without actually conveying scientific knowledge.
Hence, the lack of datasets that are based on sound definitions of science-relatedness is a crucial obstacle for advancing research into scientific online discourse and for fairly evaluating and benchmarking existing NLP and IR methods in this context. 

In order to address these challenges, we propose \textit{SciTweets}, a publicly available dataset and annotation framework for science discourse on Twitter. 

In particular, we make the following contributions:

(1) {\bf A hierarchical definition of science-relatedness.} Through an iterative process of literature review, data exploration, expert labeling and deliberation, we devise a set of definitions of science relatedness distinguishing between tweets that convey scientific knowledge in the form of claims or scientific references and tweets with a broader relatedness to research contexts and processes (Section \ref{Corpus}).

(2) {\bf Annotation framework.} Building on our set of reusable definitions, we provide an annotation framework consisting of iteratively improved and evaluated labeling instructions and a data sampling strategy informed by heuristics and a weakly supervised classifier for ensuring a balanced set of labels. 

(3) {\bf  Ground truth dataset.} We provide a dataset of 1261 tweets, labeled using our annotation framework by four expert annotators who each labeled the whole set, reaching Fleiss \(\kappa\) inter-annotator agreements between 0.61 and 0.66. All data is made publicly available under a CC-BY
Creative Commons license.

(4)  {\bf Baseline classification models. } Demonstrating the utility of our dataset and definitions, we train a baseline classifier achieving approx. 89\% F1 in distinguishing science from non-science-related tweets and 78 \% F1 in detecting science claims, references and otherwise related tweets (macro average in all cases).

\section{Constructing the SciTweets Corpus}\label{Corpus}
This section describes the annotation framework, sampling strategy, the annotation process and the resulting \textit{SciTweets} dataset. 

\subsection{Category Definitions \& Annotation Framework}
\label{subsec:framework-classes}
\begin{figure}[htbp]
\includegraphics[width=0.9\columnwidth]{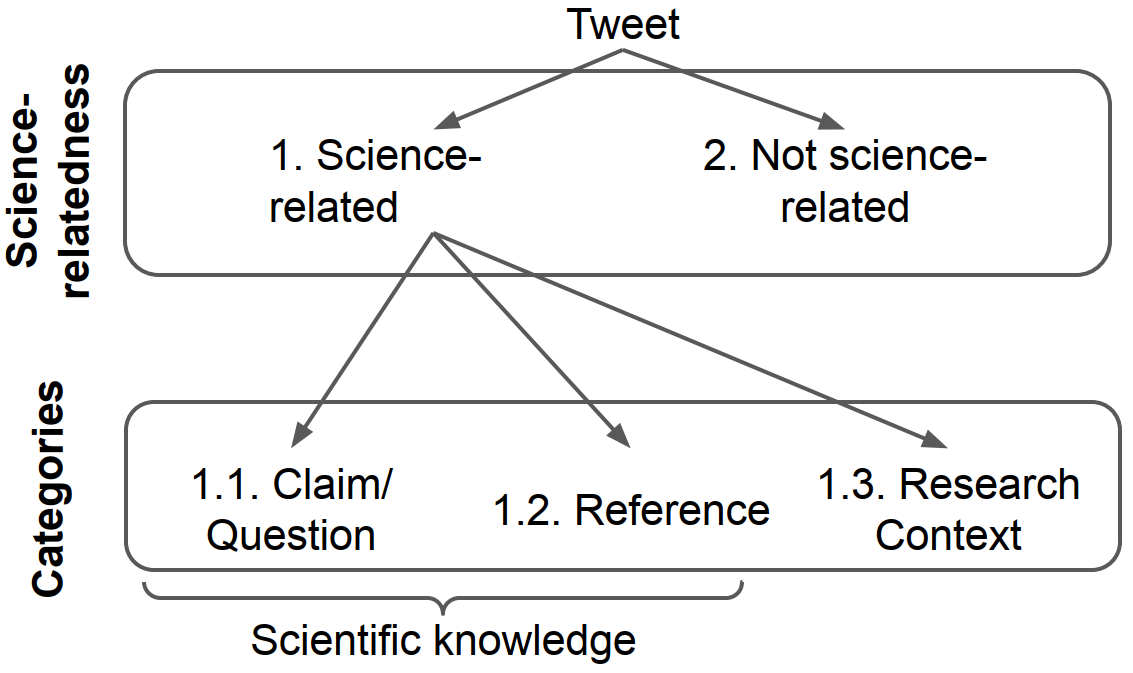}
\caption{Categories of science-relatedness}
\label{fig:subim1}
\end{figure}

Given the lack of robust definitions of science-relatedness, we followed an iterative process of data exploration, literature review and preliminary labeling rounds. We started by selecting and observing samples of science-related texts coming from science-related datasets \citep{saakyan_covid-fact_2021, wadden_fact_2020,smeros_sciclops_2021, srba_monant_2022,diggelmann_climate-fever_2021,  jansen2016extracting, Pinto2019WhatDR} and reviewing related definitions together with researchers from various disciplines. We then manually classified them into categories, and held intermediate annotation rounds with new samples to test the agreement across categories. We then identified difficult examples that had high interannotator disagreement and updated categories and annotation guidelines accordingly. In total, we held two intermediate annotation rounds with three to four annotators to improve the robustness of the categories. Definitions and labeling guidelines were considered final and ready for annotating actual ground truth data (see Section \ref{subsec:sampling-strategy}) after we obtained a satisfactory inter-annotator agreement and they facilitated an exhaustive labeling of all tweets. Categories and their definitions are described in full here\footnote{Code and data available at: https://github.com/AI-4-Sci/SciTweets}, depicted in Figure \ref{fig:subim1} and summarised below.


\noindent\textbf{Category 1 - Science-related:} Texts that fall under at least one of the following categories:
\begin{itemize}[leftmargin=0.5cm]
    \item[] \textbf{Category 1.1 - Scientific knowledge (scientifically verifiable claims):} Does the text include a claim or a question that could be scientifically verified? (see Tweet 1 in Table \ref{tab:sod_examples})
    \item[] \textbf{Category 1.2 - Reference to scientific knowledge:} Does the text include at least one reference to scientific knowledge? References can either be direct, e.g., DOI, title of a paper or indirect, e.g., a link to an article that includes a direct reference (see Tweet 2 in Table \ref{tab:sod_examples}). 
    \item[] \textbf{Category 1.3 - Related to scientific research in general:} Does the text mention a scientific research context (e.g., mention of a scientist, scientific research efforts, research findings)? (see Tweet 3 in Table \ref{tab:sod_examples})
\end{itemize}
\vspace{-\topsep}
\noindent\textbf{Category 2 - Not science-related:} Texts that don't fall under either of the 3 previous categories. (see Tweet 5 in Table \ref{tab:sod_examples})

One of the main findings from our intermediate annotation rounds was that science-relatedness cannot be defined through the presence of specific entities (e.g., "COVID-19", "vaccine") or specific domains (e.g., medicine, biology, climate) as done by several related works \citep{smeros_sciclops_2021,wadden_fact_2020}, given that the notion of scientific entity itself is ambiguous and therefore ill-defined, and hence, may lead to both false-positives and false-negatives. The reason for that is that science-relatedness is not an inherent attribute that an entity or a domain can have or not have, but rather a volatile attribute that strongly depends on the context. For instance, the word "blood" is a scientific entity in the sentence "More money is put into research efforts trying to create artificial blood", but is not a scientific entity in the sentence "He's so good at playing the guitar, it's like it's in his blood!". Therefore, for our Category 1.1 which is about texts containing scientific claims or questions, we applied the criterium of scientific verifiability, which we define as the "possibility of being verified in a document created by scientists (e.g., a scientific paper or statistics from a research institution), or verified in a document that could in theory be created by scientists, regardless of how hard that verification might be". 

Category 1.1 is crucial to distinguishing discourse that carries scientific knowledge (see Tweet 1 in Table \ref{tab:sod_examples}) from discourse that is just related to science in general, thereby enabling important tasks such as scientific claim retrieval, claim verification and claim linking. Category 1.2 (see Tweet 2 in Table \ref{tab:sod_examples}) is crucial for research that aims at understanding the role different sources play in online science discourse and the impact of science in various sources (scientific journals, online news articles, preprints, blogs). Together, Categories 1.1 and 1.2 are crucial to identifying online discourse that carries scientific knowledge in order to facilitate research into the evolution of scientific discourse in online environments. Category 1.3 is important for distinguishing discourse that does not carry scientific knowledge or a reference to scientific knowledge but nonetheless clearly mentions a scientific research context (see Tweet 3 in Table \ref{tab:sod_examples}). Tweets in that category are able to facilitate research into public perception of discourse about science and the scientific process rather than actual scientific insights. 
These three subcategories are not mutually exclusive, e.g., Tweet 4 in Table \ref{tab:sod_examples} belongs to Categories 1.1, 1.2 and 1.3. We also introduce additional \textit{Confidence Score}, \textit{Compound Claim} and \textit{Irony} labels.

\subsection{Data and Sampling Strategy}
\label{subsec:sampling-strategy}
To create our expert-annotated \textit{SciTweets} dataset, we sample tweets from the full text archive underlying TweetsKB \citep{tweetscov19}, a public knowledge graph containing metadata of more than 2 billion English tweets created from archiving 10 billion raw tweets through the 1\% Twitter API stream between February 2013 to December 2020. We extract URLs from the tweets text and resolve shortened URLs for all tweets prior to April 2018, since later ones are already extracted and resolved in the corpus.

Preliminary data exploration has shown that the percentage of science-related tweets is very small, where random sampling would surface tweets dominated by negative cases. Hence, we do not sample tweets randomly but aim to ensure a more balanced ratio of science-related and unrelated tweets. Further, we aim at including hard negative examples (e.g., “My second shot of COVID-19 vaccine gave me headache.”) instead of tweets that are obviously unrelated (e.g., “I like pop music.”). Using this approach, labeling efforts are steered towards potentially relevant cases rather than towards obviously unrelated tweets that can be obtained with high precision through random sampling combined with minimal labeling or simple heuristics. We deploy a two-stage annotation process.

\textbf{Sampling \& Annotation Stage 1.} First, we apply basic heuristics (see details here\footnote{https://github.com/AI-4-Sci/SciTweets}) on the Twitter corpus, to identify potentially science-related tweets. These identify tweets for Category 1.1 looking for patterns like nouns that are connected with argumentative predicates like 'cause' or 'lead to', filtered by a predefined list of scientific terms. Category 1.2 tweets are selected by filtering tweets that contain a URL with a subdomain that is included in a predefined list of ~17,500 scientific subdomains from open access repositories, science newspaper sections and science magazines (e.g., “link.springer.com“, “sciencedaily.com“). For Category 1.3, we retrieve tweets that mention terms and phrases related to scientists, the scientific research process and publications.
Applying this approach on a randomly sampled set of 5 million tweets obtains ~18,000 tweets that are likely to fall into either subcategory of Category 1. Given that these heuristics employ a strict pattern-matching, likely leading to high precision and low recall results, we expect the identified 18,000 tweets to lack diversity. To obtain more diverse candidate tweets, we finetune a BERTweet \citep{nguyen-etal-2020-bertweet} multi-label classifier on the 18,000 tweets as positive examples and a random sample of 18,000 tweets that were not identified by the heuristics as negative examples. Our intuition is that classification results from this classifier will lead to less precise but more diverse tweets in the final set.

After training, we use both the heuristics and the classifier to label the tweets in a randomly selected set of 100 K tweets, where each tweet is assigned two labels per category, i.e., one through the heuristics and one from the classifier. Assuming that classifier predictions will result in different tweets than the heuristics and to ensure diversity in the dataset to be annotated, we obtain all tweets where the labels of the heuristics and classifier differ for at least one of the three categories 1.1, 1.2 and 1.3, resulting in a set of 1046 tweets that were annotated in a first labeling step (Section \ref{Annotation}). 

\textbf{Sampling \& Annotation Stage 2.} These 1046 expert annotations obtained in the first annotation stage are used to train a new multi-label classifier for the three science-relatedness subcategories (see details in Section \ref{Classifier}). After applying the classifier to a new set of 100K randomly selected tweets, we obtain the 100 tweets with the highest confidence from the resulting predictions for each category, i.e., 300 tweets in total. After filtering out duplicates, this results in 215 additional tweets, that were again annotated by expert annotators \ref{Annotation}), resulting in a total set of 1261 expert-annotated tweets. 
\subsection{Annotation \& Quality Assurance}\label{Annotation}
All tweets are labeled by the same four annotators, including the two main co-authors of this paper as well as a PhD student and a bachelor's student, both from the field of Computer Science. Before annotating, we held individual training sessions with the annotators in which we examined examples and counterexamples of each category of the labeling framework to ensure the labeling task was understood correctly. Cases of weak disagreement, i.e., where the label of one annotator differs to the labels of the three remaining annotators, were consolidated using a majority vote. For tweets from the second annotation stage, cases of high disagreement, i.e., where both labels \textit{Yes} and \textit{No} were each selected by two annotators, were resolved in a discussion between all four annotators, whereas we did not resolve high disagreements for tweets from the first annotation stage. We measured the inter-annotator agreement by computing Fleiss Kappa \(\kappa\) \citep{Fleiss1971MeasuringNS} with the labels of all four annotators to evaluate the annotation quality. Agreement scores of 0.61, 0.63 and 0.66 for categories 1.1, 1.2 and 1.3 averaging a score of 0.63 are comparable to results on similar tasks\footnote{\citet{thorne2018fever} achieved a fleiss score of 0.68 on a fact verification task with five annotators and \citet{alam2020fighting} achieved a score of 0.75 on a task of determining whether a tweet contains a verifiable factual claim.}. Since we estimate our task to be more difficult than those tasks and because the fleiss score has been shown to be different based on the number of annotators \citep{Fleiss1971MeasuringNS}, we estimate our results to be comparable to the mentioned papers and thus to be encouraging. 

\subsection{Statistics}\label{AnnotationStats}
The \textit{SciTweets} dataset consists of 1261 human-annotated tweets including the labels of the individual annotators, as well as the consolidated ground-truth label for each category. Table \ref{tab:stats1} shows the distributions of the ground-truth labels with a high ratio (31.88\%) of science-related tweets (Category 1), where the consolidated label to at least one science-relatedness subcategory is \textit{Yes}, and a balanced distribution of subcategories ranging from 15.65\% for Category 1.2 to 23.82\% for Category 1.1.

\begin{table}[htbp]
  \caption{Distribution of the ground-truth labels}
  \label{tab:stats1}
  \begin{tabular}{c@{\hspace{0.4em}}r@{\hspace{0.4em}}|r@{\hspace{0.4em}}r@{\hspace{0.4em}}r@{\hspace{0.4em}}}
    \toprule
    Labels  &Category 1 &Category 1.1 &Category 1.2 &Category 1.3\\
    \midrule
    Yes & 402 (31.88\%)&  283 (23.82\%)    &   190 (15.65\%) &   259 (21.32\%)\\
    No  & 859 (68.12\%)&  905 (76.18\%)    &   1024 (84.35\%)&   956 (78.68\%)\\
  \bottomrule
\end{tabular}
\end{table}

\vspace{-1.5em}
\section{Classification of Science-relatedness}\label{Classifier}
We evaluate a single multi-label classifier for both the binary task of classifying if a tweet is science-related as well as the multi-label task of assigning one or more subcategories of science-relatedness to a tweet. Experimenting with different base models showed that SciBERT \citep{beltagy-etal-2019-scibert} provides superior performance on the tasks. To evaluate the multi-label classifier on the binary task we map the classifier's multi-label predictions to a binary prediction, i.e., the classifier predicts a tweet to be science-related if it assigns at least one of the three subcategories 1.1, 1.2 or 1.3. Table \ref{tab:classifier1} shows the performance of the classifier for both tasks applying 10-fold cross validation using all \textit{SciTweets} tweets without high disagreements. As expected, the classifier performs better on the binary task, because a false positive prediction for categories 1.1, 1.2, and 1.3 could still be a true positive prediction for the binary task. Given the high ratio of science-related tweets in \textit{SciTweets} compared to TweetsKB, the precision for both tasks is expected to be lower when performed on a random sample of TweetsKB, because of the increase of false positives. Hence, to get a more representative performance estimate, we train a new multilabel classifier on the 1046 tweets from annotation stage 1 and set a prediction threshold for each subcategory, so that the classifier makes only 100 positive predictions per subcategory out of 100K tweets. Table \ref{tab:classifier2} shows the precision for these positive predictions (215 tweets, labeled in the second annotation stage).
\begin{table}[htpb]
  \caption{Classifier performance (binary and multilabel tasks)}
  \label{tab:classifier1}
  \begin{tabular}{llrrr}
    \toprule
    Task & Category  & Precision & Recall    & F1\\
    \midrule
    \multirow{2}{*}{binary} & 1 - Science-related   &   84.70      &   83.99    &   84.34  \\
                            & 2 - Not Science-related         &   92.67      &   93.03    &   92.85  \\
    \midrule
    \multirow{3}{*}{multi}  & 1.1 - Scientific Claim             &   75.00      &   81.18    &   77.97\\
                            & 1.2 - Reference        &   76.19      &   77.01    &   76.60  \\
                            & 1.3 - Research Context           &   81.06      &   79.65    &   80.35  \\
  \bottomrule
\end{tabular}
\end{table}
\begin{table}[htpb]
\caption{Classifier Performance for multilabel task}
\label{tab:classifier2}
\begin{tabular}{lrrr}
 \toprule
 Metric  &Category 1.1 &Category 1.2     &Category 1.3\\
 \midrule
 Precision@100  &    85.00\%      &   74.00\%           &   86.00\%       \\
\bottomrule
\end{tabular}
\end{table}

\section{Conclusion}
Resources and claims related to science contribute significantly to online discourse, in particular with respect to emerging topics of high societal importance, such as climate change or COVID-19. The understanding of science-related online discourse can help prevent the spread of science-related misinformation through the help of computational methods that facilitate research across various disciplines. Foundations of such research and methods are sound definitions of science relatedness and reliable and publicly available datasets that enable the advancement and evaluation of methods dealing with downstream tasks such as (scientific) claims detection, retrieval, classification or verification. In this paper, we propose a hierarchical definition of science-relatedness underlying an annotation framework for science-related tweets that forms the basis of \textit{SciTweets}, an unprecedented annotated ground-truth dataset for science discourse on Twitter. Based on this data, we train a baseline classifier for detection of science-relatedness, showing promising initial results. Whereas \textit{SciTweets} is a comparably small corpus, it provides a high-quality ground truth for testing models,  where the heuristics used as part of our sampling methodology open directions for future work by obtaining large-scale weakly labeled training data for training models.

\begin{acks}
This work is supported by the AI4Sci grant, co-funded by MESRI (France, grant UM-211745), BMBF (Germany, grant 01IS21086), and the French National Research Agency (ANR).
\end{acks}

\bibliographystyle{ACM-Reference-Format}
\bibliography{references.bib}


\begin{thebibliography}{31}


\ifx \showCODEN    \undefined \def \showCODEN     #1{\unskip}     \fi
\ifx \showDOI      \undefined \def \showDOI       #1{#1}\fi
\ifx \showISBNx    \undefined \def \showISBNx     #1{\unskip}     \fi
\ifx \showISBNxiii \undefined \def \showISBNxiii  #1{\unskip}     \fi
\ifx \showISSN     \undefined \def \showISSN      #1{\unskip}     \fi
\ifx \showLCCN     \undefined \def \showLCCN      #1{\unskip}     \fi
\ifx \shownote     \undefined \def \shownote      #1{#1}          \fi
\ifx \showarticletitle \undefined \def \showarticletitle #1{#1}   \fi
\ifx \showURL      \undefined \def \showURL       {\relax}        \fi
\providecommand\bibfield[2]{#2}
\providecommand\bibinfo[2]{#2}
\providecommand\natexlab[1]{#1}
\providecommand\showeprint[2][]{arXiv:#2}

\bibitem[\protect\citeauthoryear{Alam, Shaar, Dalvi, Sajjad, Nikolov, Mubarak,
  Martino, Abdelali, Durrani, Darwish, et~al\mbox{.}}{Alam
  et~al\mbox{.}}{2020}]%
        {alam2020fighting}
\bibfield{author}{\bibinfo{person}{Firoj Alam}, \bibinfo{person}{Shaden Shaar},
  \bibinfo{person}{Fahim Dalvi}, \bibinfo{person}{Hassan Sajjad},
  \bibinfo{person}{Alex Nikolov}, \bibinfo{person}{Hamdy Mubarak},
  \bibinfo{person}{Giovanni Da~San Martino}, \bibinfo{person}{Ahmed Abdelali},
  \bibinfo{person}{Nadir Durrani}, \bibinfo{person}{Kareem Darwish},
  {et~al\mbox{.}}} \bibinfo{year}{2020}\natexlab{}.
\newblock \showarticletitle{Fighting the COVID-19 infodemic: modeling the
  perspective of journalists, fact-checkers, social media platforms, policy
  makers, and the society}.
\newblock \bibinfo{journal}{\emph{arXiv preprint arXiv:2005.00033}}
  (\bibinfo{year}{2020}).
\newblock


\bibitem[\protect\citeauthoryear{Allcott and Gentzkow}{Allcott and
  Gentzkow}{2017}]%
        {AlcotttFakeNews}
\bibfield{author}{\bibinfo{person}{Hunt Allcott} {and} \bibinfo{person}{Matthew
  Gentzkow}.} \bibinfo{year}{2017}\natexlab{}.
\newblock \showarticletitle{Social Media and Fake News in the 2016 Election}.
\newblock \bibinfo{journal}{\emph{Journal of Economic Perspectives}}
  \bibinfo{volume}{31}, \bibinfo{number}{2} (\bibinfo{date}{May}
  \bibinfo{year}{2017}), \bibinfo{pages}{211--36}.
\newblock
\urldef\tempurl%
\url{https://doi.org/10.1257/jep.31.2.211}
\showDOI{\tempurl}


\bibitem[\protect\citeauthoryear{Banerjee, Kelkar, Logan, Majhail, and
  Pemmaraju}{Banerjee et~al\mbox{.}}{2021}]%
        {banerjee2021democratization}
\bibfield{author}{\bibinfo{person}{Rahul Banerjee}, \bibinfo{person}{Amar~H
  Kelkar}, \bibinfo{person}{Aaron~C Logan}, \bibinfo{person}{Navneet~S
  Majhail}, {and} \bibinfo{person}{Naveen Pemmaraju}.}
  \bibinfo{year}{2021}\natexlab{}.
\newblock \showarticletitle{The democratization of scientific conferences:
  Twitter in the era of COVID-19 and beyond}.
\newblock \bibinfo{journal}{\emph{Current hematologic malignancy reports}}
  \bibinfo{volume}{16}, \bibinfo{number}{2} (\bibinfo{year}{2021}),
  \bibinfo{pages}{132--139}.
\newblock


\bibitem[\protect\citeauthoryear{Beltagy, Lo, and Cohan}{Beltagy
  et~al\mbox{.}}{2019}]%
        {beltagy-etal-2019-scibert}
\bibfield{author}{\bibinfo{person}{Iz Beltagy}, \bibinfo{person}{Kyle Lo},
  {and} \bibinfo{person}{Arman Cohan}.} \bibinfo{year}{2019}\natexlab{}.
\newblock \showarticletitle{{S}ci{BERT}: A Pretrained Language Model for
  Scientific Text}. In \bibinfo{booktitle}{\emph{Proceedings of the 2019
  Conference on Empirical Methods in Natural Language Processing and the 9th
  International Joint Conference on Natural Language Processing
  (EMNLP-IJCNLP)}}. \bibinfo{publisher}{Association for Computational
  Linguistics}, \bibinfo{address}{Hong Kong, China},
  \bibinfo{pages}{3615--3620}.
\newblock
\urldef\tempurl%
\url{https://doi.org/10.18653/v1/D19-1371}
\showDOI{\tempurl}


\bibitem[\protect\citeauthoryear{Carlson and Harris}{Carlson and
  Harris}{2020}]%
        {carlson2020quantifying}
\bibfield{author}{\bibinfo{person}{Jedidiah Carlson} {and}
  \bibinfo{person}{Kelley Harris}.} \bibinfo{year}{2020}\natexlab{}.
\newblock \showarticletitle{Quantifying and contextualizing the impact of
  bioRxiv preprints through automated social media audience segmentation}.
\newblock \bibinfo{journal}{\emph{PLoS Biology}} \bibinfo{volume}{18},
  \bibinfo{number}{9} (\bibinfo{year}{2020}), \bibinfo{pages}{e3000860}.
\newblock


\bibitem[\protect\citeauthoryear{D{\'\i}az-Faes, Bowman, and
  Costas}{D{\'\i}az-Faes et~al\mbox{.}}{2019}]%
        {diaz2019towards}
\bibfield{author}{\bibinfo{person}{Adri{\'a}n~A D{\'\i}az-Faes},
  \bibinfo{person}{Timothy~D Bowman}, {and} \bibinfo{person}{Rodrigo Costas}.}
  \bibinfo{year}{2019}\natexlab{}.
\newblock \showarticletitle{Towards a second generation of ‘social media
  metrics’: Characterizing Twitter communities of attention around science}.
\newblock \bibinfo{journal}{\emph{PloS one}} \bibinfo{volume}{14},
  \bibinfo{number}{5} (\bibinfo{year}{2019}), \bibinfo{pages}{e0216408}.
\newblock


\bibitem[\protect\citeauthoryear{Diggelmann, Boyd-Graber, Bulian, Ciaramita,
  and Leippold}{Diggelmann et~al\mbox{.}}{2021}]%
        {diggelmann_climate-fever_2021}
\bibfield{author}{\bibinfo{person}{Thomas Diggelmann}, \bibinfo{person}{Jordan
  Boyd-Graber}, \bibinfo{person}{Jannis Bulian}, \bibinfo{person}{Massimiliano
  Ciaramita}, {and} \bibinfo{person}{Markus Leippold}.}
  \bibinfo{year}{2021}\natexlab{}.
\newblock \showarticletitle{{CLIMATE}-{FEVER}: {A} {Dataset} for {Verification}
  of {Real}-{World} {Climate} {Claims}}.
\newblock \bibinfo{journal}{\emph{arXiv:2012.00614 [cs]}} (\bibinfo{date}{Jan.}
  \bibinfo{year}{2021}).
\newblock
\urldef\tempurl%
\url{http://arxiv.org/abs/2012.00614}
\showURL{%
\tempurl}
\newblock
\shownote{arXiv: 2012.00614.}


\bibitem[\protect\citeauthoryear{Dimitrov, Baran, Fafalios, Yu, Zhu, Zloch, and
  Dietze}{Dimitrov et~al\mbox{.}}{2020}]%
        {tweetscov19}
\bibfield{author}{\bibinfo{person}{Dimitar Dimitrov}, \bibinfo{person}{Erdal
  Baran}, \bibinfo{person}{Pavlos Fafalios}, \bibinfo{person}{Ran Yu},
  \bibinfo{person}{Xiaofei Zhu}, \bibinfo{person}{Matth\"{a}us Zloch}, {and}
  \bibinfo{person}{Stefan Dietze}.} \bibinfo{year}{2020}\natexlab{}.
\newblock \showarticletitle{TweetsCOV19 - A Knowledge Base of Semantically
  Annotated Tweets about the COVID-19 Pandemic}. In
  \bibinfo{booktitle}{\emph{Proceedings of the 29th ACM International
  Conference on Information \& Knowledge Management}} (Virtual Event, Ireland)
  \emph{(\bibinfo{series}{CIKM '20})}. \bibinfo{publisher}{Association for
  Computing Machinery}, \bibinfo{address}{New York, NY, USA},
  \bibinfo{pages}{2991–2998}.
\newblock
\showISBNx{9781450368599}
\urldef\tempurl%
\url{https://doi.org/10.1145/3340531.3412765}
\showDOI{\tempurl}


\bibitem[\protect\citeauthoryear{Fleiss}{Fleiss}{1971}]%
        {Fleiss1971MeasuringNS}
\bibfield{author}{\bibinfo{person}{Joseph~L. Fleiss}.}
  \bibinfo{year}{1971}\natexlab{}.
\newblock \showarticletitle{Measuring nominal scale agreement among many
  raters.}
\newblock \bibinfo{journal}{\emph{Psychological Bulletin}}
  \bibinfo{volume}{76} (\bibinfo{year}{1971}), \bibinfo{pages}{378--382}.
\newblock


\bibitem[\protect\citeauthoryear{Garimella, Morales, Gionis, and
  Mathioudakis}{Garimella et~al\mbox{.}}{2018}]%
        {quanticontroversy}
\bibfield{author}{\bibinfo{person}{Kiran Garimella}, \bibinfo{person}{Gianmarco
  De~Francisci Morales}, \bibinfo{person}{Aristides Gionis}, {and}
  \bibinfo{person}{Michael Mathioudakis}.} \bibinfo{year}{2018}\natexlab{}.
\newblock \showarticletitle{Quantifying Controversy on Social Media}.
\newblock \bibinfo{journal}{\emph{Trans. Soc. Comput.}} \bibinfo{volume}{1},
  \bibinfo{number}{1}, Article \bibinfo{articleno}{3} (\bibinfo{date}{jan}
  \bibinfo{year}{2018}), \bibinfo{numpages}{27}~pages.
\newblock
\showISSN{2469-7818}
\urldef\tempurl%
\url{https://doi.org/10.1145/3140565}
\showDOI{\tempurl}


\bibitem[\protect\citeauthoryear{Hassan, Zhang, Arslan, Caraballo, Jimenez,
  Gawsane, Hasan, Joseph, Kulkarni, Nayak, Sable, Li, and Tremayne}{Hassan
  et~al\mbox{.}}{2017}]%
        {hassan_claimbuster_2017}
\bibfield{author}{\bibinfo{person}{Naeemul Hassan}, \bibinfo{person}{Gensheng
  Zhang}, \bibinfo{person}{Fatma Arslan}, \bibinfo{person}{Josue Caraballo},
  \bibinfo{person}{Damian Jimenez}, \bibinfo{person}{Siddhant Gawsane},
  \bibinfo{person}{Shohedul Hasan}, \bibinfo{person}{Minumol Joseph},
  \bibinfo{person}{Aaditya Kulkarni}, \bibinfo{person}{Anil~Kumar Nayak},
  \bibinfo{person}{Vikas Sable}, \bibinfo{person}{Chengkai Li}, {and}
  \bibinfo{person}{Mark Tremayne}.} \bibinfo{year}{2017}\natexlab{}.
\newblock \showarticletitle{{ClaimBuster}: the first-ever end-to-end
  fact-checking system}.
\newblock \bibinfo{journal}{\emph{Proceedings of the VLDB Endowment}}
  \bibinfo{volume}{10}, \bibinfo{number}{12} (\bibinfo{date}{Aug.}
  \bibinfo{year}{2017}), \bibinfo{pages}{1945--1948}.
\newblock
\showISSN{2150-8097}
\urldef\tempurl%
\url{https://doi.org/10.14778/3137765.3137815}
\showDOI{\tempurl}


\bibitem[\protect\citeauthoryear{Haunschild, Bornmann, Potnis, and
  Tahamtan}{Haunschild et~al\mbox{.}}{2021}]%
        {haunschild2021investigating}
\bibfield{author}{\bibinfo{person}{Robin Haunschild}, \bibinfo{person}{Lutz
  Bornmann}, \bibinfo{person}{Devendra Potnis}, {and} \bibinfo{person}{Iman
  Tahamtan}.} \bibinfo{year}{2021}\natexlab{}.
\newblock \showarticletitle{Investigating dissemination of scientific
  information on Twitter: A study of topic networks in opioid publications}.
\newblock \bibinfo{journal}{\emph{Quantitative Science Studies}}
  (\bibinfo{year}{2021}), \bibinfo{pages}{1--56}.
\newblock


\bibitem[\protect\citeauthoryear{Iyengar and Massey}{Iyengar and
  Massey}{2019}]%
        {doi:10.1073/pnas.1805868115}
\bibfield{author}{\bibinfo{person}{Shanto Iyengar} {and}
  \bibinfo{person}{Douglas~S. Massey}.} \bibinfo{year}{2019}\natexlab{}.
\newblock \showarticletitle{Scientific communication in a post-truth society}.
\newblock \bibinfo{journal}{\emph{Proceedings of the National Academy of
  Sciences}} \bibinfo{volume}{116}, \bibinfo{number}{16}
  (\bibinfo{year}{2019}), \bibinfo{pages}{7656--7661}.
\newblock
\urldef\tempurl%
\url{https://doi.org/10.1073/pnas.1805868115}
\showDOI{\tempurl}
\showeprint{https://www.pnas.org/doi/pdf/10.1073/pnas.1805868115}


\bibitem[\protect\citeauthoryear{Jansen and Kuhn}{Jansen and Kuhn}{2016}]%
        {jansen2016extracting}
\bibfield{author}{\bibinfo{person}{Tom Jansen} {and} \bibinfo{person}{Tobias
  Kuhn}.} \bibinfo{year}{2016}\natexlab{}.
\newblock \showarticletitle{Extracting core claims from scientific articles}.
  In \bibinfo{booktitle}{\emph{Benelux Conference on Artificial Intelligence}}.
  Springer, \bibinfo{pages}{32--46}.
\newblock


\bibitem[\protect\citeauthoryear{Kreps and Kriner}{Kreps and Kriner}{2020}]%
        {doi:10.1126/sciadv.abd4563}
\bibfield{author}{\bibinfo{person}{S.~E. Kreps} {and} \bibinfo{person}{D.~L.
  Kriner}.} \bibinfo{year}{2020}\natexlab{}.
\newblock \showarticletitle{Model uncertainty, political contestation, and
  public trust in science: Evidence from the COVID-19 pandemic}.
\newblock \bibinfo{journal}{\emph{Science Advances}} \bibinfo{volume}{6},
  \bibinfo{number}{43} (\bibinfo{year}{2020}), \bibinfo{pages}{eabd4563}.
\newblock
\urldef\tempurl%
\url{https://doi.org/10.1126/sciadv.abd4563}
\showDOI{\tempurl}
\showeprint{https://www.science.org/doi/pdf/10.1126/sciadv.abd4563}


\bibitem[\protect\citeauthoryear{Li, Akin, Su, Brossard, Xenos, and
  Scheufele}{Li et~al\mbox{.}}{2016}]%
        {li2016tweeting}
\bibfield{author}{\bibinfo{person}{Nan Li}, \bibinfo{person}{Heather Akin},
  \bibinfo{person}{Leona Yi-Fan Su}, \bibinfo{person}{Dominique Brossard},
  \bibinfo{person}{Michael~A Xenos}, {and} \bibinfo{person}{Dietram
  Scheufele}.} \bibinfo{year}{2016}\natexlab{}.
\newblock \showarticletitle{Tweeting disaster: An analysis of online discourse
  about nuclear power in the wake of the Fukushima Daiichi nuclear accident}.
\newblock \bibinfo{journal}{\emph{Journal of Science Communication}}
  \bibinfo{volume}{15}, \bibinfo{number}{5} (\bibinfo{year}{2016}),
  \bibinfo{pages}{A02}.
\newblock


\bibitem[\protect\citeauthoryear{Moukarzel, Rehm, Del~Fresno, and
  Daly}{Moukarzel et~al\mbox{.}}{2020}]%
        {moukarzel2020diffusing}
\bibfield{author}{\bibinfo{person}{Sara Moukarzel}, \bibinfo{person}{Martin
  Rehm}, \bibinfo{person}{Miguel Del~Fresno}, {and} \bibinfo{person}{Alan~J
  Daly}.} \bibinfo{year}{2020}\natexlab{}.
\newblock \showarticletitle{Diffusing science through social networks: The case
  of breastfeeding communication on Twitter}.
\newblock \bibinfo{journal}{\emph{PloS one}} \bibinfo{volume}{15},
  \bibinfo{number}{8} (\bibinfo{year}{2020}), \bibinfo{pages}{e0237471}.
\newblock


\bibitem[\protect\citeauthoryear{Nakov, Barr\'{o}n-Cede\~{n}o, Da~San~Martino,
  Alam, Stru\ss{}, Mandl, M\'{\i}guez, Caselli, Kutlu, Zaghouani, Li, Shaar,
  Shahi, Mubarak, Nikolov, Babulkov, Kartal, and Beltr\'{a}n}{Nakov
  et~al\mbox{.}}{2022}]%
        {clefcheckthat2022}
\bibfield{author}{\bibinfo{person}{Preslav Nakov}, \bibinfo{person}{Alberto
  Barr\'{o}n-Cede\~{n}o}, \bibinfo{person}{Giovanni Da~San~Martino},
  \bibinfo{person}{Firoj Alam}, \bibinfo{person}{Julia~Maria Stru\ss{}},
  \bibinfo{person}{Thomas Mandl}, \bibinfo{person}{Rub\'{e}n M\'{\i}guez},
  \bibinfo{person}{Tommaso Caselli}, \bibinfo{person}{Mucahid Kutlu},
  \bibinfo{person}{Wajdi Zaghouani}, \bibinfo{person}{Chengkai Li},
  \bibinfo{person}{Shaden Shaar}, \bibinfo{person}{Gautam~Kishore Shahi},
  \bibinfo{person}{Hamdy Mubarak}, \bibinfo{person}{Alex Nikolov},
  \bibinfo{person}{Nikolay Babulkov}, \bibinfo{person}{Yavuz~Selim Kartal},
  {and} \bibinfo{person}{Javier Beltr\'{a}n}.} \bibinfo{year}{2022}\natexlab{}.
\newblock \showarticletitle{The CLEF-2022 CheckThat! Lab on Fighting the
  COVID-19 Infodemic and Fake News Detection}. In
  \bibinfo{booktitle}{\emph{Advances in Information Retrieval: 44th European
  Conference on IR Research, ECIR 2022, Stavanger, Norway, April 10–14, 2022,
  Proceedings, Part II}} (Stavanger, Norway).
  \bibinfo{publisher}{Springer-Verlag}, \bibinfo{address}{Berlin, Heidelberg},
  \bibinfo{pages}{416–428}.
\newblock
\showISBNx{978-3-030-99738-0}
\urldef\tempurl%
\url{https://doi.org/10.1007/978-3-030-99739-7_52}
\showDOI{\tempurl}


\bibitem[\protect\citeauthoryear{Nakov, Corney, Hasanain, Alam, Elsayed,
  Barrón-Cedeño, Papotti, Shaar, and Martino}{Nakov et~al\mbox{.}}{2021}]%
        {nakov_automated_2021}
\bibfield{author}{\bibinfo{person}{Preslav Nakov}, \bibinfo{person}{David
  Corney}, \bibinfo{person}{Maram Hasanain}, \bibinfo{person}{Firoj Alam},
  \bibinfo{person}{Tamer Elsayed}, \bibinfo{person}{Alberto Barrón-Cedeño},
  \bibinfo{person}{Paolo Papotti}, \bibinfo{person}{Shaden Shaar}, {and}
  \bibinfo{person}{Giovanni Da~San Martino}.} \bibinfo{year}{2021}\natexlab{}.
\newblock \showarticletitle{Automated {Fact}-{Checking} for {Assisting} {Human}
  {Fact}-{Checkers}}.
\newblock \bibinfo{journal}{\emph{arXiv:2103.07769 [cs]}} (\bibinfo{date}{May}
  \bibinfo{year}{2021}).
\newblock
\urldef\tempurl%
\url{http://arxiv.org/abs/2103.07769}
\showURL{%
\tempurl}
\newblock
\shownote{arXiv: 2103.07769.}


\bibitem[\protect\citeauthoryear{Nguyen, Vu, and Tuan~Nguyen}{Nguyen
  et~al\mbox{.}}{2020}]%
        {nguyen-etal-2020-bertweet}
\bibfield{author}{\bibinfo{person}{Dat~Quoc Nguyen}, \bibinfo{person}{Thanh
  Vu}, {and} \bibinfo{person}{Anh Tuan~Nguyen}.}
  \bibinfo{year}{2020}\natexlab{}.
\newblock \showarticletitle{{BERT}weet: A pre-trained language model for
  {E}nglish Tweets}. In \bibinfo{booktitle}{\emph{Proceedings of the 2020
  Conference on Empirical Methods in Natural Language Processing: System
  Demonstrations}}. \bibinfo{publisher}{Association for Computational
  Linguistics}, \bibinfo{address}{Online}, \bibinfo{pages}{9--14}.
\newblock
\urldef\tempurl%
\url{https://doi.org/10.18653/v1/2020.emnlp-demos.2}
\showDOI{\tempurl}


\bibitem[\protect\citeauthoryear{Pinto, Wawrzinek, and Balke}{Pinto
  et~al\mbox{.}}{2019}]%
        {Pinto2019WhatDR}
\bibfield{author}{\bibinfo{person}{Jos{\'e} Mar{\'i}a~Gonz{\'a}lez Pinto},
  \bibinfo{person}{Janus Wawrzinek}, {and} \bibinfo{person}{Wolf-Tilo Balke}.}
  \bibinfo{year}{2019}\natexlab{}.
\newblock \showarticletitle{What Drives Research Efforts? Find Scientific
  Claims that Count!}
\newblock \bibinfo{journal}{\emph{2019 ACM/IEEE Joint Conference on Digital
  Libraries (JCDL)}} (\bibinfo{year}{2019}), \bibinfo{pages}{217--226}.
\newblock


\bibitem[\protect\citeauthoryear{Romanou, Smeros, Castillo, and Aberer}{Romanou
  et~al\mbox{.}}{2020}]%
        {Scilens}
\bibfield{author}{\bibinfo{person}{Angelika Romanou},
  \bibinfo{person}{Panayiotis Smeros}, \bibinfo{person}{Carlos Castillo}, {and}
  \bibinfo{person}{Karl Aberer}.} \bibinfo{year}{2020}\natexlab{}.
\newblock \showarticletitle{SciLens News Platform: {A} System for Real-Time
  Evaluation of News Articles}.
\newblock \bibinfo{journal}{\emph{Proc. {VLDB} Endow.}} \bibinfo{volume}{13},
  \bibinfo{number}{12} (\bibinfo{year}{2020}), \bibinfo{pages}{2969--2972}.
\newblock
\urldef\tempurl%
\url{http://www.vldb.org/pvldb/vol13/p2969-romanou.pdf}
\showURL{%
\tempurl}


\bibitem[\protect\citeauthoryear{Saakyan, Chakrabarty, and Muresan}{Saakyan
  et~al\mbox{.}}{2021}]%
        {saakyan_covid-fact_2021}
\bibfield{author}{\bibinfo{person}{Arkadiy Saakyan}, \bibinfo{person}{Tuhin
  Chakrabarty}, {and} \bibinfo{person}{Smaranda Muresan}.}
  \bibinfo{year}{2021}\natexlab{}.
\newblock \showarticletitle{{COVID}-{Fact}: {Fact} {Extraction} and
  {Verification} of {Real}-{World} {Claims} on {COVID}-19 {Pandemic}}. In
  \bibinfo{booktitle}{\emph{Proceedings of the 59th {Annual} {Meeting} of the
  {Association} for {Computational} {Linguistics} and the 11th {International}
  {Joint} {Conference} on {Natural} {Language} {Processing} ({Volume} 1: {Long}
  {Papers})}}. \bibinfo{publisher}{Association for Computational Linguistics},
  \bibinfo{address}{Online}, \bibinfo{pages}{2116--2129}.
\newblock
\urldef\tempurl%
\url{https://doi.org/10.18653/v1/2021.acl-long.165}
\showDOI{\tempurl}


\bibitem[\protect\citeauthoryear{Smeros, Castillo, and Aberer}{Smeros
  et~al\mbox{.}}{2021}]%
        {smeros_sciclops_2021}
\bibfield{author}{\bibinfo{person}{Panayiotis Smeros}, \bibinfo{person}{Carlos
  Castillo}, {and} \bibinfo{person}{Karl Aberer}.}
  \bibinfo{year}{2021}\natexlab{}.
\newblock \showarticletitle{{SciClops}: {Detecting} and {Contextualizing}
  {Scientific} {Claims} for {Assisting} {Manual} {Fact}-{Checking}}.
\newblock \bibinfo{journal}{\emph{Proceedings of the 30th ACM International
  Conference on Information \& Knowledge Management}} (\bibinfo{date}{Oct.}
  \bibinfo{year}{2021}), \bibinfo{pages}{1692--1702}.
\newblock
\urldef\tempurl%
\url{https://doi.org/10.1145/3459637.3482475}
\showDOI{\tempurl}
\newblock
\shownote{arXiv: 2110.13090.}


\bibitem[\protect\citeauthoryear{Srba, Pecher, Tomlein, Moro, Stefancova,
  Simko, and Bielikova}{Srba et~al\mbox{.}}{2022}]%
        {srba_monant_2022}
\bibfield{author}{\bibinfo{person}{Ivan Srba}, \bibinfo{person}{Branislav
  Pecher}, \bibinfo{person}{Matus Tomlein}, \bibinfo{person}{Robert Moro},
  \bibinfo{person}{Elena Stefancova}, \bibinfo{person}{Jakub Simko}, {and}
  \bibinfo{person}{Maria Bielikova}.} \bibinfo{year}{2022}\natexlab{}.
\newblock \showarticletitle{Monant {Medical} {Misinformation} {Dataset}:
  {Mapping} {Articles} to {Fact}-{Checked} {Claims}}.
\newblock \bibinfo{journal}{\emph{arXiv:2204.12294 [cs]}}
  (\bibinfo{date}{April} \bibinfo{year}{2022}).
\newblock
\urldef\tempurl%
\url{https://doi.org/10.1145/3477495.3531726}
\showDOI{\tempurl}
\newblock
\shownote{arXiv: 2204.12294.}


\bibitem[\protect\citeauthoryear{Thorne, Vlachos, Christodoulopoulos, and
  Mittal}{Thorne et~al\mbox{.}}{2018}]%
        {thorne2018fever}
\bibfield{author}{\bibinfo{person}{James Thorne}, \bibinfo{person}{Andreas
  Vlachos}, \bibinfo{person}{Christos Christodoulopoulos}, {and}
  \bibinfo{person}{Arpit Mittal}.} \bibinfo{year}{2018}\natexlab{}.
\newblock \showarticletitle{Fever: a large-scale dataset for fact extraction
  and verification}.
\newblock \bibinfo{journal}{\emph{arXiv preprint arXiv:1803.05355}}
  (\bibinfo{year}{2018}).
\newblock


\bibitem[\protect\citeauthoryear{van Schalkwyk and Dudek}{van Schalkwyk and
  Dudek}{2022}]%
        {doi:10.1177/09636625221077392}
\bibfield{author}{\bibinfo{person}{Francois van Schalkwyk} {and}
  \bibinfo{person}{Jonathan Dudek}.} \bibinfo{year}{2022}\natexlab{}.
\newblock \showarticletitle{Reporting preprints in the media during the
  COVID-19 pandemic}.
\newblock \bibinfo{journal}{\emph{Public Understanding of Science}}
  \bibinfo{volume}{31}, \bibinfo{number}{5} (\bibinfo{year}{2022}),
  \bibinfo{pages}{608--616}.
\newblock
\urldef\tempurl%
\url{https://doi.org/10.1177/09636625221077392}
\showDOI{\tempurl}
\showeprint{https://doi.org/10.1177/09636625221077392}
\newblock
\shownote{PMID: 35196912.}


\bibitem[\protect\citeauthoryear{Vosoughi, Roy, and Aral}{Vosoughi
  et~al\mbox{.}}{2018}]%
        {vosoughi}
\bibfield{author}{\bibinfo{person}{Soroush Vosoughi}, \bibinfo{person}{Deb
  Roy}, {and} \bibinfo{person}{Sinan Aral}.} \bibinfo{year}{2018}\natexlab{}.
\newblock \showarticletitle{The spread of true and false news online}.
\newblock \bibinfo{journal}{\emph{Science}} \bibinfo{volume}{359},
  \bibinfo{number}{6380} (\bibinfo{year}{2018}), \bibinfo{pages}{1146--1151}.
\newblock
\urldef\tempurl%
\url{https://doi.org/10.1126/science.aap9559}
\showDOI{\tempurl}
\showeprint{https://www.science.org/doi/pdf/10.1126/science.aap9559}


\bibitem[\protect\citeauthoryear{Wadden, Lin, Lo, Wang, van Zuylen, Cohan, and
  Hajishirzi}{Wadden et~al\mbox{.}}{2020}]%
        {wadden_fact_2020}
\bibfield{author}{\bibinfo{person}{David Wadden}, \bibinfo{person}{Shanchuan
  Lin}, \bibinfo{person}{Kyle Lo}, \bibinfo{person}{Lucy~Lu Wang},
  \bibinfo{person}{Madeleine van Zuylen}, \bibinfo{person}{Arman Cohan}, {and}
  \bibinfo{person}{Hannaneh Hajishirzi}.} \bibinfo{year}{2020}\natexlab{}.
\newblock \showarticletitle{Fact or {Fiction}: {Verifying} {Scientific}
  {Claims}}.
\newblock \bibinfo{journal}{\emph{arXiv:2004.14974 [cs]}} (\bibinfo{date}{Oct.}
  \bibinfo{year}{2020}).
\newblock
\urldef\tempurl%
\url{http://arxiv.org/abs/2004.14974}
\showURL{%
\tempurl}
\newblock
\shownote{arXiv: 2004.14974.}


\bibitem[\protect\citeauthoryear{Walter, L{\"o}rcher, and
  Br{\"u}ggemann}{Walter et~al\mbox{.}}{2019}]%
        {walter2019scientific}
\bibfield{author}{\bibinfo{person}{Stefanie Walter}, \bibinfo{person}{Ines
  L{\"o}rcher}, {and} \bibinfo{person}{Michael Br{\"u}ggemann}.}
  \bibinfo{year}{2019}\natexlab{}.
\newblock \showarticletitle{Scientific networks on Twitter: Analyzing
  scientists’ interactions in the climate change debate}.
\newblock \bibinfo{journal}{\emph{Public Understanding of Science}}
  \bibinfo{volume}{28}, \bibinfo{number}{6} (\bibinfo{year}{2019}),
  \bibinfo{pages}{696--712}.
\newblock


\bibitem[\protect\citeauthoryear{Wright, Wadden, Lo, Kuehl, Cohan, Augenstein,
  and Wang}{Wright et~al\mbox{.}}{2022}]%
        {wright-etal-2022-generating}
\bibfield{author}{\bibinfo{person}{Dustin Wright}, \bibinfo{person}{David
  Wadden}, \bibinfo{person}{Kyle Lo}, \bibinfo{person}{Bailey Kuehl},
  \bibinfo{person}{Arman Cohan}, \bibinfo{person}{Isabelle Augenstein}, {and}
  \bibinfo{person}{Lucy Wang}.} \bibinfo{year}{2022}\natexlab{}.
\newblock \showarticletitle{Generating Scientific Claims for Zero-Shot
  Scientific Fact Checking}. In \bibinfo{booktitle}{\emph{Proceedings of the
  60th Annual Meeting of the Association for Computational Linguistics (Volume
  1: Long Papers)}}. \bibinfo{publisher}{Association for Computational
  Linguistics}, \bibinfo{address}{Dublin, Ireland},
  \bibinfo{pages}{2448--2460}.
\newblock
\urldef\tempurl%
\url{https://aclanthology.org/2022.acl-long.175}
\showURL{%
\tempurl}


\end{thebibliography}



\end{document}